\documentclass[conference]{IEEEtran}
\IEEEoverridecommandlockouts

\usepackage{cite}
\usepackage{amsmath,amssymb,amsfonts}
\usepackage{algorithmic}
\usepackage{graphicx}
\usepackage{textcomp}
\usepackage{xcolor}
\def\BibTeX{{\rm B\kern-.05em{\sc i\kern-.025em b}\kern-.08em
    T\kern-.1667em\lower.7ex\hbox{E}\kern-.125emX}}

\usepackage{booktabs} 
\usepackage{multirow}
\usepackage[super]{nth}
\usepackage{tikz}

\newcommand*\circledB[1]{\tikz[baseline=(char.base)]{
            \node[shape=circle,fill,inner sep=0.2pt] (char) {\textcolor{white}{#1}};}}

\usepackage{soul}
\usepackage{enumitem}
\usepackage{algorithmic}
\usepackage{algorithm}
\usepackage{array}
\newcolumntype{?}{!{\vrule width 1.5pt}}
\usepackage{fancyhdr}
\fancypagestyle{firstpage}{
  \fancyhf{}
  \chead{\small To appear at the 2023 International Conference on Automation, Robotics and Applications (ICARA), February 2023, Abu Dhabi, UAE.}
  \cfoot{\thepage}
}

\usepackage{setspace}

\begin{document}

%%%%%%%%%%%%%%%%%%%%%%%%%%%%%%%%%%%%%%%%%%%%%%%%%%%%%%%%
\title{Mantis: Enabling Energy-Efficient Autonomous \\ Mobile Agents with Spiking Neural Networks
\vspace{-0.2cm}}

\author{\IEEEauthorblockN{Rachmad Vidya Wicaksana Putra\IEEEauthorrefmark{1}, Muhammad Shafique\IEEEauthorrefmark{2}}
\IEEEauthorblockA{\IEEEauthorrefmark{1}\textit{Institute of Computer Engineering}, \textit{Technische Universit\"at Wien (TU Wien)}, Vienna, Austria\\
\IEEEauthorrefmark{2}\textit{Division of Engineering}, \textit{New York University Abu Dhabi (NYUAD)}, Abu Dhabi, United Arab Emirates\\
Email: rachmad.putra@tuwien.ac.at, muhammad.shafique@nyu.edu}
\vspace{-0.9cm}}

\maketitle
\pagestyle{plain}
\thispagestyle{firstpage}

%%%%%%%%%%%%%%%%%%%%%%%%%%%%%%%%%%%%%%%%%%%%%%%%%%%%%%%%
\begin{spacing}{0.96}
\begin{abstract}
Autonomous mobile agents such as unmanned aerial vehicles (UAVs) and mobile robots have shown huge potential for improving human productivity.
These mobile agents require low power/energy consumption to have a long lifespan since they are usually powered by batteries. 
These agents also need to adapt to changing/dynamic environments, especially when deployed in far or dangerous locations, thus requiring efficient online learning capabilities.
These requirements can be fulfilled by employing Spiking Neural Networks (SNNs) since SNNs offer low power/energy consumption due to sparse computations and efficient online learning due to bio-inspired learning mechanisms. 
However, a methodology is still required to employ appropriate SNN models on autonomous mobile agents. 
Towards this, we propose a Mantis methodology to systematically employ SNNs on autonomous mobile agents to enable energy-efficient processing and adaptive capabilities in dynamic environments. 
The key ideas of our Mantis include the optimization of SNN operations, the employment of a bio-plausible online learning mechanism, and the SNN model selection. 
The experimental results demonstrate that our methodology maintains high accuracy with a significantly smaller memory footprint and energy consumption (i.e., 3.32x memory reduction and 2.9x energy saving for an SNN model with 8-bit weights) compared to the baseline network with 32-bit weights. 
In this manner, our Mantis enables the employment of SNNs for resource- and energy-constrained mobile agents. 
\end{abstract}

\end{spacing}

%%%%%%%%%%%%%%%%%%%%%%%%%%%%%%%%%%%%%%%%%%%%%%%%%%%%%%%%
\begin{spacing}{0.96}
\vspace{-0.1cm}
\section{Introduction}
\label{Sec_Intro}

\begin{figure}[t]
\centering
\includegraphics[width=0.86\linewidth]{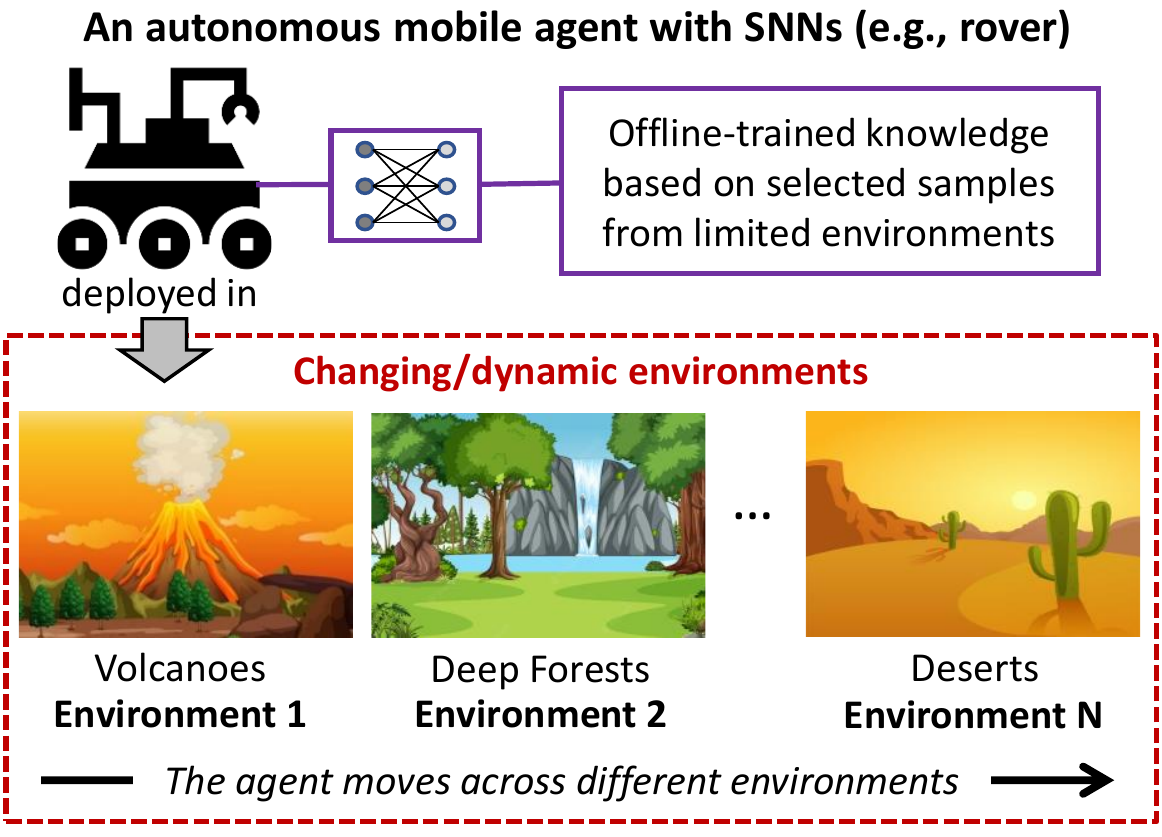}
\vspace{-0.3cm}
\caption{Autonomous mobile agents (e.g., rover) may encounter different environments from the one used in the offline training, therefore the offline-trained knowledge may be obsolete during the operational time.}
\label{Fig_DynamicEnv}
\vspace{-0.4cm}
\end{figure}

Autonomous mobile agents, such as unmanned aerial vehicles (UAVs) and mobile robots, have emerged as effective solutions to explore and study objects and/or environments that are physically too far or unreachable by humans (e.g., volcanoes, forests, and even other planets)~\cite{Ref_Putra_lpSpikeCon_IJCNN22}.
The ability to explore environments is important for autonomous agents, and one of the prominent techniques is object recognition through images or videos~\cite{Ref_Bonnevie_LongTermExplore_ICARA23}.
Therefore, mobile agents require \textit{low power/energy consumption} to have a long lifespan since they are usually powered by batteries. 
Furthermore, these agents also require \textit{efficient online learning} mechanisms to adapt to changing/dynamic environments~\cite{Ref_Putra_SpikeDyn_DAC21}, as shown in Fig.~\ref{Fig_DynamicEnv}. 

\begin{figure}[h]
\centering
\includegraphics[width=0.86\linewidth]{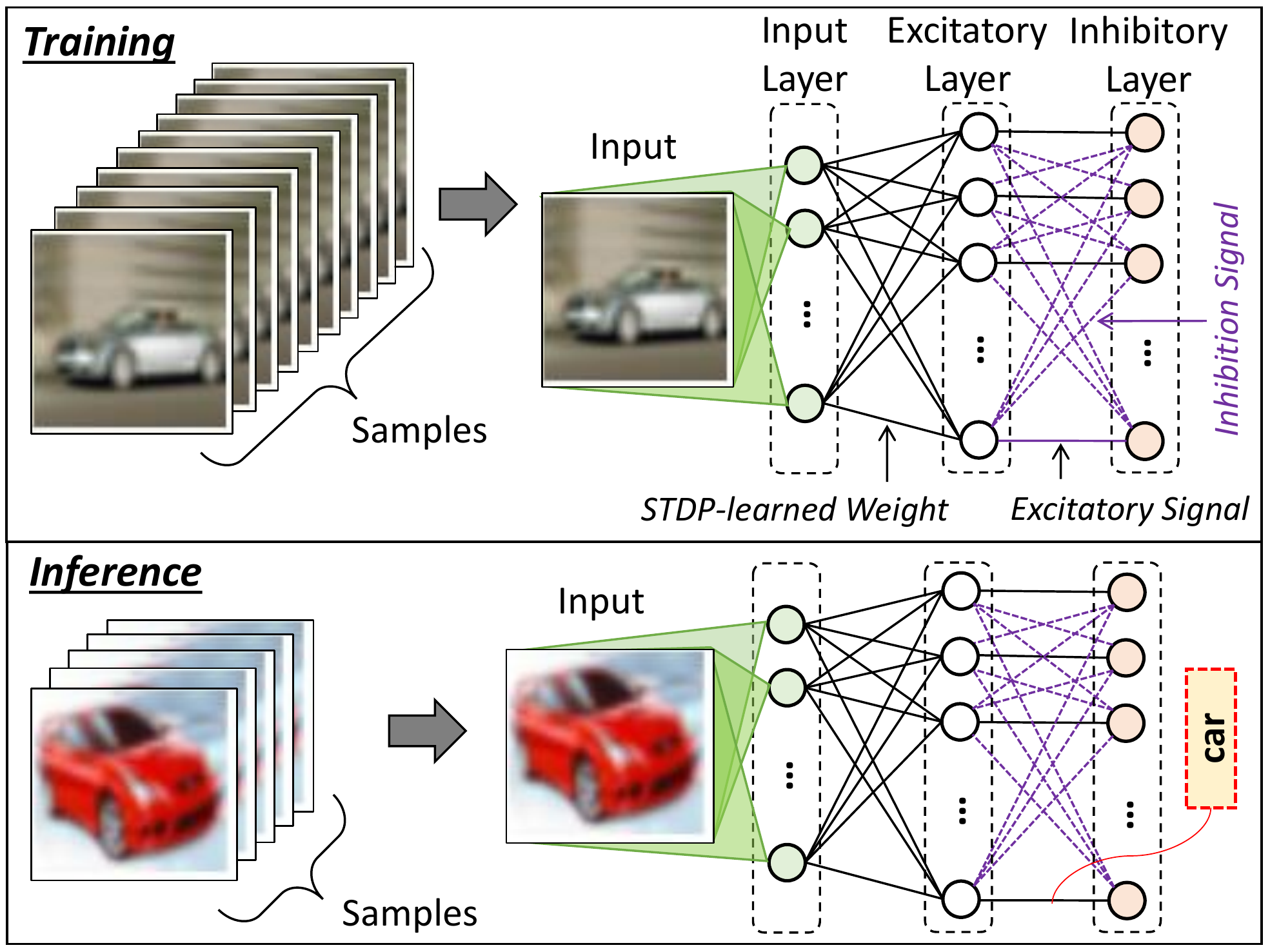}
\vspace{-0.3cm}
\caption{Training and inference phases for SNNs. We consider the fully-connected network as it offers high accuracy under various unsupervised STDP-based learning rules~\cite{Ref_Diehl_STDPmnist_FNCOM15}. The unsupervised learning rules are important for enabling efficient online learning scenarios, since these rules can learn from unlabeled data that are directly obtained from operational environments.}
\label{Fig_SNNs}
\vspace{-0.5cm}
\end{figure}

Spiking Neural Networks (SNNs) are considered the suitable solution for fulfilling the above requirements posed by autonomous mobile agents. 
The reason is that, SNNs can offer high accuracy with ultra-low power/energy consumption~\cite{Ref_Pfeiffer_DLSNN_FNINS18, Ref_Tavanaei_DLSNN_Neunet18,Ref_Diehl_STDPmnist_FNCOM15, Ref_Hazan_SOMSNN_IJCNN18, Ref_Saunders_STDPpatch_IJCNN18, Ref_Saunders_LCSNN_NeuNet19, Ref_Kaiser_DECOLLE_FNINS20, Ref_Putra_FSpiNN_TCAD20} and unsupervised learning mechanisms which are suitable for online learning scenarios~\cite{Ref_Putra_lpSpikeCon_IJCNN22}\cite{Ref_Putra_SpikeDyn_DAC21}\cite{Ref_Putra_SoftSNN_DAC22} due to their bio-plausible spike-based operations and learning rule, e.g., Spike-Timing-Dependent Plasticity (STDP).
Recent trends suggest that large SNN models (i.e., a large number of synapses and neurons) are frequently used in state-of-the-art works since they can achieve higher accuracy than the smaller ones due to their capabilities for recognizing more features~\cite{Ref_Shafique_EdgeAI_ICCAD21}\cite{Ref_Putra_ReSpawn_ICCAD21}. 
For instance, a fully-connected SNN (shown in Fig.~\ref{Fig_SNNs}) with 32-bit floating-point format (FP32) consumes $\sim$200MB of memory and achieves 92\% accuracy on the MNIST dataset, while a smaller model with $\sim$1MB of memory achieves 75\% accuracy~\cite{Ref_Putra_FSpiNN_TCAD20}. 
Therefore, state-of-the-art SNNs usually have a large number of parameters (e.g., weights and neuron parameters) that occupy a large memory footprint. 
This leads to intensive memory accesses, which dominate the energy consumption of neural network-based computations~\cite{Ref_Putra_SparkXD_DAC21, Ref_Krithivasan_SpikeBundle_ISLPED19, Ref_Putra_EnforceSNN_FNINS22}, thereby hindering the deployment of SNNs on the resource- and energy-constrained autonomous mobile agents.

\textbf{Targeted Problem:}
\textit{How can we employ SNNs to meet the memory and energy requirements of autonomous mobile agents while keeping their high accuracy.} 

%%%%%%%%%
\subsection{Limitations of State-of-the-Art}
\label{Sec_Prelim_SOTA}

\begin{figure}[hbtp]
\centering
\includegraphics[width=0.9\linewidth]{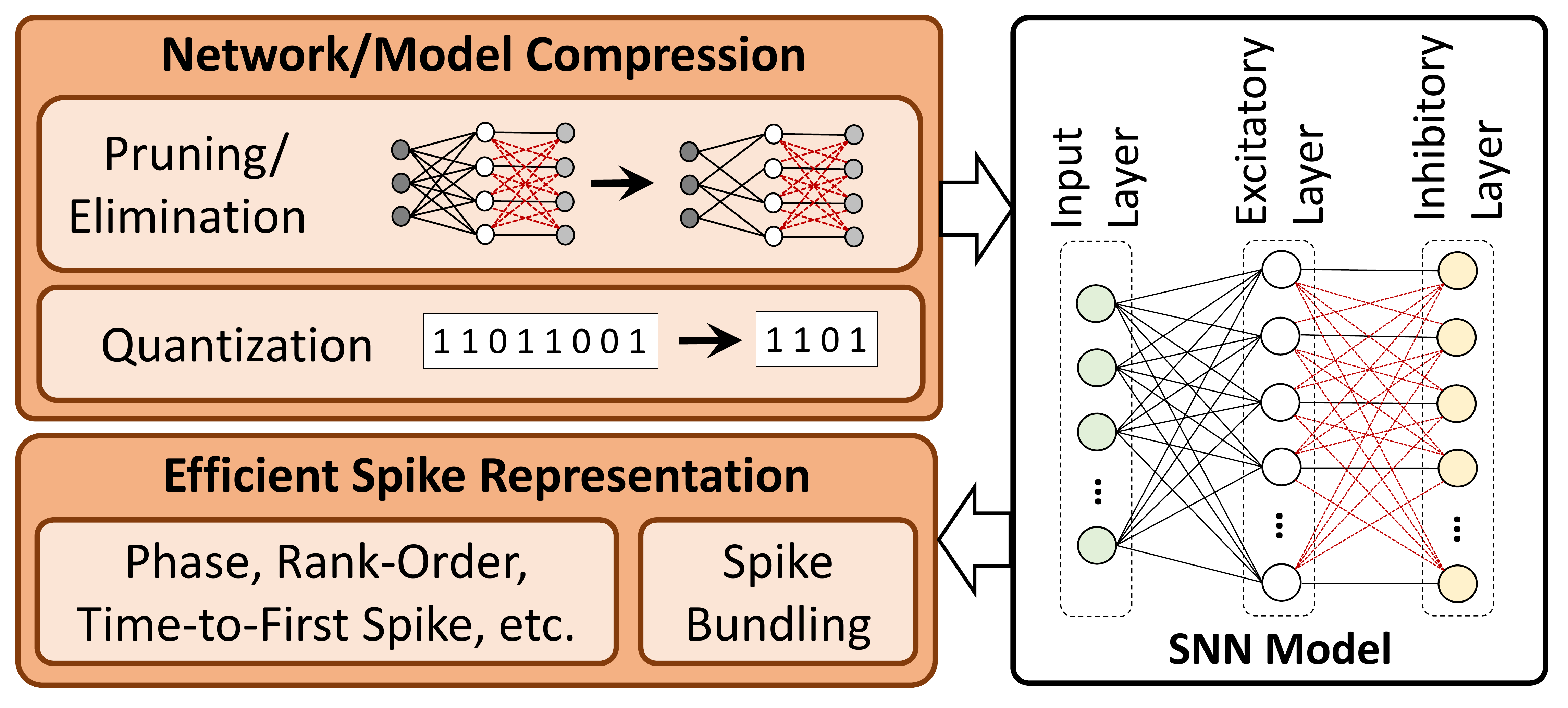}
\vspace{-0.3cm}
\caption{The existing techniques for optimizing the memory footprint of SNNs that leads to the reduction of energy consumption.}
\label{Fig_OverviewTech}
\vspace{-0.4cm}
\end{figure}

Fig.~\ref{Fig_OverviewTech} highlights the state-of-the-art techniques for optimizing the memory footprint of SNNs that lead to energy saving. 
Works of~\cite{Ref_Rathi_PruneQuantizeSNN_TCAD18}\cite{Ref_Hu_QuantSTDPSNN_NeCo21} perform quantization and/or weight pruning to compress the SNN model for inference. 
However, they still require a huge number of (inhibitory) neurons to effectively perform inference and suffer from accuracy degradation as compared to the baseline model (i.e., dense and non-quantized model).  
Other works employ different spike coding schemes to optimize the memory footprint for storing the spike information~\cite{Ref_Kayser_PhaseCoding_Neuron09}\cite{Ref_Park_T2FSNN_DAC20}. 
The work of~\cite{Ref_Krithivasan_SpikeBundle_ISLPED19} reduces the memory accesses by bundling a sequence of spikes into a single spike. 
However, these spike representation techniques do not optimize SNN parameters, which dominate the memory footprint and access energy.
Other works mainly focus on improving the accuracy at the cost of additional computations, which leads to high memory and energy requirements~\cite{Ref_Hazan_SOMSNN_IJCNN18}\cite{Ref_Saunders_LCSNN_NeuNet19}\cite{Ref_Panda_ASP_JETCAS18, Ref_Srinivasan_EnhPlast_IJCNN17, Ref_Hazan_LMSNN_AMAI19}. 
For instance, the work of~\cite{Ref_Srinivasan_EnhPlast_IJCNN17} improves STDP-based learning by ensuring that the update is essential. 
However, it needs a huge number of neurons to perform inference effectively.
Therefore, these techniques only offer partial benefits, i.e., either memory reduction, energy saving, or accuracy improvement, thereby making them unsuitable for autonomous mobile agents, which require all the above benefits. 
Furthermore, to adapt to different operational environments, such mobile agents\footnote{For conciseness, we use terms ``mobile agents" to represent ``autonomous mobile agents".} also require an efficient online learning mechanism so that they can perform training at run time for updating the knowledge. 
However, such training scenarios need to avoid \textit{catastrophic forgetting}, i.e., the system quickly forgets the previously learned knowledge after learning the new ones~\cite{Ref_Putra_SpikeDyn_DAC21}.
Recent works have tried to address this issue through the employment of additional neurons~\cite{Ref_Allred_ForcedFiring_IJCNN16}\cite{Ref_Allred_CFN_FNINS20}, adaptive learning rates~\cite{Ref_Putra_lpSpikeCon_IJCNN22}\cite{Ref_Putra_SpikeDyn_DAC21}\cite{Ref_Panda_ASP_JETCAS18}, and adaptive neuron dynamics~\cite{Ref_Putra_lpSpikeCon_IJCNN22}.
 
The above discussion shows that the existing techniques are developed mainly to improve the performance and efficiency of SNN models, however how mobile agents can benefit from these techniques have not been studied. 
Therefore, there is a significant need for \textit{a design methodology that systematically employs SNNs on autonomous mobile agents while meeting the given memory and energy budgets}. 

%%%%%%%%%
\subsection{Contributions of This Work}
\label{Sec_Intro_Contrib}

We propose \textit{\textbf{Mantis methodology} to enable energy-efficient autonomous \underline{M}obile \underline{a}ge\underline{nt}s by employ\underline{i}ng \underline{s}piking neural networks}.
Our Mantis employs the following steps.
\begin{itemize}[leftmargin=*]
    \item \textbf{Employment of SNN Optimizations (Section~\ref{Sec_Method_ModelOpt}):}
    This step aims at producing an SNN model that can maintain accuracy as compared to the baseline model but with smaller model size. 
    We employ optimization techniques that reduce SNN operations and quantize SNN parameters.
    \item \textbf{Employment of Enhanced Learning Rule (Section~\ref{Sec_Method_Learning}):} 
    We employ an STDP-based learning rule mechanism that addresses catastrophic forgetting issues by employing an adaptive learning rate, weight decay, and balanced threshold potential, while considering weight quantization. 
    \item \textbf{SNN Model Selection Algorithm (Section~\ref{Sec_Method_ModelSel}):}
    We select the SNN model that achieves acceptable accuracy and meets the given memory and energy budgets as the solution for the mobile agents.
    To do this, we develop a model selection algorithm and a memory and energy estimator to find the suitable model for the given use-case scenarios. 
\end{itemize}

%%%%%%%%%%%%%%%%%%%%%%%%%%%%%%%%%%%%%%%%%%%%%%%%%%%%%%%%
\section{Mantis Methodology}

%%%%%%%%%
\subsection{Overview}
\label{Sec_Method_Overview}

\begin{figure}[hbtp]
\vspace{-0.2cm}
\centering
\includegraphics[width=\linewidth]{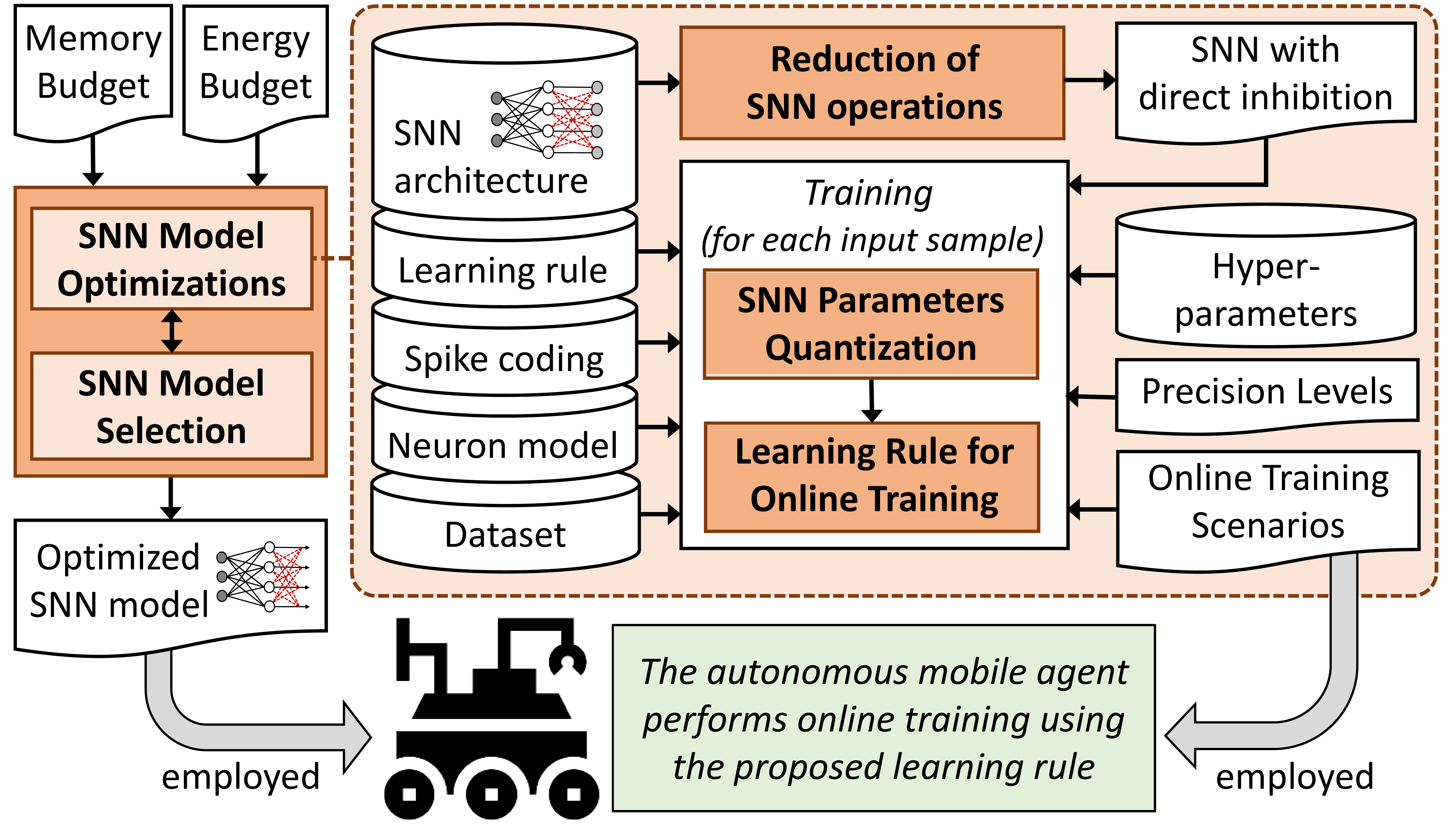}
\vspace{-0.6cm}
\caption{The Mantis methodology showing its key steps. 
}
\label{Fig_OurMethod}
\vspace{-0.3cm}
\end{figure}

Our Mantis methodology employs several key steps; see Fig.~\ref{Fig_OurMethod}. 
These steps are described in the following subsections. 

%%%%%%%%%
\subsection{SNN Model Optimizations}
\label{Sec_Method_ModelOpt}

%%%%
\subsubsection{Reduction of Operations}
\label{Sec_Method_ModelOpt_Ops}

\begin{figure}[hbtp]
%\vspace{-0.2cm}
\centering
\includegraphics[width=\linewidth]{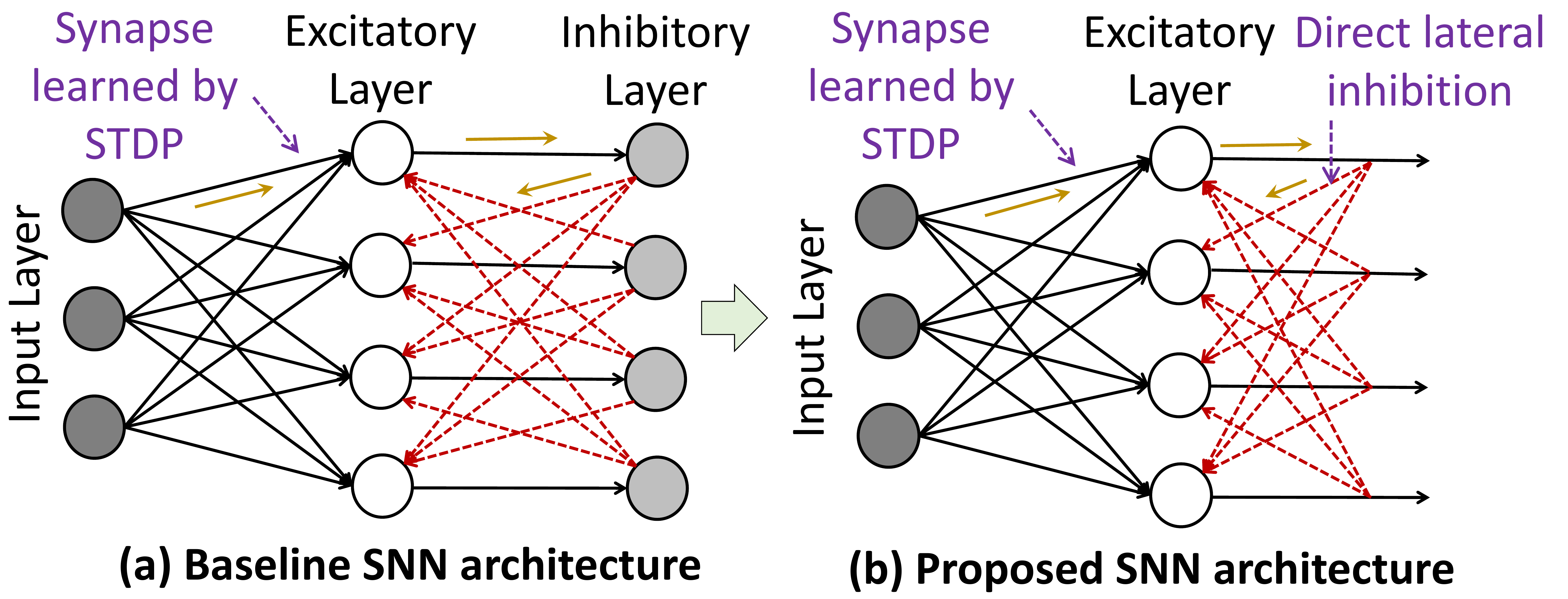}
\vspace{-0.6cm}
\caption{The proposed reduction of SNN operations; adapted from~\cite{Ref_Putra_FSpiNN_TCAD20}.}
\label{Fig_ModifiedSNN}
\vspace{-0.4cm}
\end{figure}

We observe that the inhibitory neurons require different parameter values from the excitatory ones to generate inhibitory spikes properly for effective SNN processing. 
Therefore, parameters of inhibitory neurons also need to be stored in memory.
This indicates that a large number of inhibitory neurons may lead to a large memory footprint and energy consumption. 
Towards this, \textit{we employ techniques that substitute the inhibitory neurons with lateral inhibition to decrease the number of neuron operations and parameters~\cite{Ref_Putra_FSpiNN_TCAD20}, hence curtailing the memory footprint and energy consumption for mobile agents}, as shown in Fig.~\ref{Fig_ModifiedSNN}. 
Therefore, the function of inhibitory spikes (i.e., providing competition among neurons) is replaced by spikes from excitatory neurons.

%%%%%
\subsubsection{Parameter Quantization}
\label{Sec_Method_ModelOpt_Quant}

An SNN model typically has different types of parameters (e.g., weights and neuron parameters) that need to be stored in the memory during the training and inference phases.
We observe that these parameters can be quantized to minimize the memory footprint and energy consumption.
Towards this, \textit{we employ techniques that quantize different SNN parameters (i.e., weights and neuron parameters) while exploring different precision levels~\cite{Ref_Putra_QSpiNN_IJCNN21}, thereby exploring candidates of SNN models with appropriate sizes for mobile agents}.
For the quantization scheme, we consider the quantization-aware training/in-training quantization~\cite{Ref_Krishnamoorthi_Whitepaper_arXiv18} to facilitate the online learning scenario, i.e., the quantized SNNs can perform training online to update their knowledge.
The idea of our quantization is to maximize the compression for each SNN parameter, and define the precision level based on its impact on the accuracy. 
In this work, for our case study, we quantize weights and two neuron parameters, i.e., membrane potential ($V_{mem}$) and threshold potential ($V_{th}$).

%%%%%%%%%
\subsection{Employment of the Enhanced Learning Rule}
\label{Sec_Method_Learning}

The optimized SNN model should effectively learn the input features during online training at run time. 
Towards this, \textit{we employ techniques that enhance the baseline STDP-based learning rule through adaptive learning rates, weight decay, and balanced threshold potential~\cite{Ref_Putra_SpikeDyn_DAC21}}.

\vspace{0.1cm}
%%%%%
\subsubsection{Adaptive Learning Rates}
We employ the potentiation factor ($f_p$) and depression factor ($f_d$) to enhance the learning rates for weight potentiation and depression, respectively.
The idea is that the $f_p$ should be set with a high value when the connecting synapses need to learn input features, which is indicated by the appearance of postsynaptic spikes. 
The value of $f_p$ is defined by normalizing the maximum number of postsynaptic spikes ($S_{post}$) with a spike threshold ($S_{th}$). 
Meanwhile, the $f_d$ should be set with a high value when the connecting synapses need to be weakened, which is indicated by no appearance of postsynaptic spikes.
The value of $f_d$ is defined by normalizing the maximum accumulated postsynaptic spikes ($S_{post}$) and presynaptic spikes ($S_{pre}$). 
These enhancement factors are employed in the enhanced STDP-based learning algorithm, as shown in Eq.~\ref{Eq_STDP_Improved}.
Note, $\Delta w$ is the value of weight change, $\eta_{pre}$ and $\eta_{post}$ are the learning rates for a presynaptic and postsynaptic spike, respectively. Meanwhile, $x_{pre}$ and $x_{post}$ are the presynaptic and postsynaptic traces, respectively.

\vspace{-0.3cm}
\begin{equation}
\small
\begin{split}
\Delta w = 
\begin{cases}
-f_d \cdot \eta_{pre} \cdot  x_{post} & \text{at} \; \text{depression update}\\
f_p \cdot \eta_{post} \cdot x_{pre} & \text{at} \; \text{potentiation update}
\end{cases}
\label{Eq_STDP_Improved}
\end{split}
\end{equation}
%\vspace{-0.3cm}

\vspace{0.2cm}
%%%%%
\subsubsection{Weight Decay}
We consider weight decay to remove the old and insignificant knowledge, which is indicated by a small weight value at the defined weight decay rate~$w_{decay}$.
In this way, weak synapse connections will be more disconnected over the time of training, and these synapses can be used to learn new features.
To do this, the value of $w_{decay}$ is set to be inversely proportional to the model size as a bigger model has a higher number of synapses for learning new knowledge, hence it removes the old knowledge at a slower rate than the smaller networks. 

\vspace{0.2cm}
%%%%%
\subsubsection{Balanced Threshold Potential}
We find that the threshold potential ($V_{th}$) has a role in determining if a neuron would produce spikes frequently.
If $V_{th}$ is too low, the neurons will generate spikes frequently for any inputs, hence leading the network to erase the old knowledge.
Meanwhile, if $V_{th}$ is too high, the neurons will not spike easily, hence the network will have difficulties for learning new features.
Therefore, the $V_{th}$ should be balanced so that some neurons may recognize new features while others keep the old important knowledge.

%%%%%%%%%
\subsection{Memory- and Energy-aware Model Selection}
\label{Sec_Method_ModelSel}

The compressed models from the optimization techniques need to be evaluated to select the most appropriate one that provides high accuracy and meets the memory and energy constraints. 
Towards this, \textit{we propose a model selection algorithm that consider the accuracy, memory, and energy of the investigated model}, which has the following steps.
\begin{enumerate}[leftmargin=*]
    \item First, we check if the accuracy of the investigated model is greater than or equal to ($\geq$) the lowest acceptable accuracy.
        \begin{itemize}
        \item If yes, we move to step-2.
        \item If no, the investigated model is not selected, and we can evaluate another model and move to step-1 again.
        \end{itemize}
     \item We check if the memory footprint of the investigated model is less than or equal to ($\leq$) the memory budget. We estimate the memory footprint using Eq.~\ref{Eq_Memory}. 
        \begin{equation}
        \small
        \begin{split}
        M = Mw + Mn = Nw \cdot Bw + \sum_k Nn_k \cdot Bn_k
        \end{split}
        \label{Eq_Memory}
        \end{equation}
        The memory footprint ($M$) is the total size of the weights ($Mw$) and neuron parameters ($Mn$).  
        $Mw$ equals to the number of weights ($Nw$) multiplied by the corresponding bitwidth ($Bw$).
        Meanwhile, $Mn$ equals to the number of neuron parameter-\textit{k} ($Nn_k$) multiplied by the corresponding bitwidth ($Bn_k$).
        \begin{itemize}
            \item If yes, we move to step-3.
            \item If no, the investigated model is not selected, and we can evaluate another model and move to step-1 again.
        \end{itemize}
    \item We check if the energy consumption of the investigated model is less than or equal to ($\leq$) the memory budget. We estimate the energy consumption using Eq.~\ref{Eq_Energy}. 
        \begin{equation}
        \small
        \begin{split}
        E = E_1 \cdot N
        \end{split}
        \label{Eq_Energy}
        \end{equation}
        The total energy ($E$) is estimated by leveraging the energy consumption for processing a single sample ($E_1$) measured directly from the platform and the number of samples that will be processed ($N$). 
        \begin{itemize}
            \item If yes, the investigated model is selected as a solution candidate, and we move to step-4.
            \item If no, the investigated model is not selected, then we can evaluate another model and move to step-1 again.
        \end{itemize}
    \item If we have multiple candidates, then we select the one with the best performance in a specific priority as the solution (e.g., either accuracy, memory, or energy). Otherwise, we only have a single candidate as the solution.  
\end{enumerate}

%%%%%%%%%%%%%%%%%%%%%%%%%%%%%%%%%%%%%%%%%%%%%%%%%%%%%%%%
\section{Evaluation Methodology}
\label{Sec_EvalMethod}

We use PyTorch-based simulations~\cite{Ref_Hazan_BindsNET_FNINF18} for evaluating SNN accuracy, which run on GPUs (i.e., Nvidia GeForce RTX 2080 Ti for GPGPUs and Nvidia Jetson Nano for embedded GPUs).
Evaluations with embedded GPUs aim at investigating the applicability of our Mantis methodology for autonomous agents, as such platforms are suitable for energy-constrained mobile applications. 
We consider the fully-connected SNN architecture (see Fig.~\ref{Fig_ModifiedSNN}) with 400 excitatory neurons, while employing the rate coding. 
We employ the MNIST and Fashion MNIST datasets as workloads. 
For comparison, we consider the SNN design with pair excitatory-inhibitory neurons~\cite{Ref_Diehl_STDPmnist_FNCOM15} (see Fig.~\ref{Fig_ModifiedSNN}a) and enhanced pair-wise STDP~\cite{Ref_Putra_SpikeDyn_DAC21} as the baseline. 
To estimate the memory, we leverage the number of weights and neuron parameters as well as their bitwidth.
Then, we leverage the operational power and the simulation time to estimate energy consumption.

%%%%%%%%%%%%%%%%%%%%%%%%%%%%%%%%%%%%%%%%%%%%%%%%%%%%%%%%
\section{Results and Discussion}
\label{Sec_Results}

%%%%%%%%%
\subsection{Maintaining Accuracy}
\label{Sec_Results_Accuracy}

\begin{figure*}[hbtp]
\centering
\includegraphics[width=0.87\linewidth]{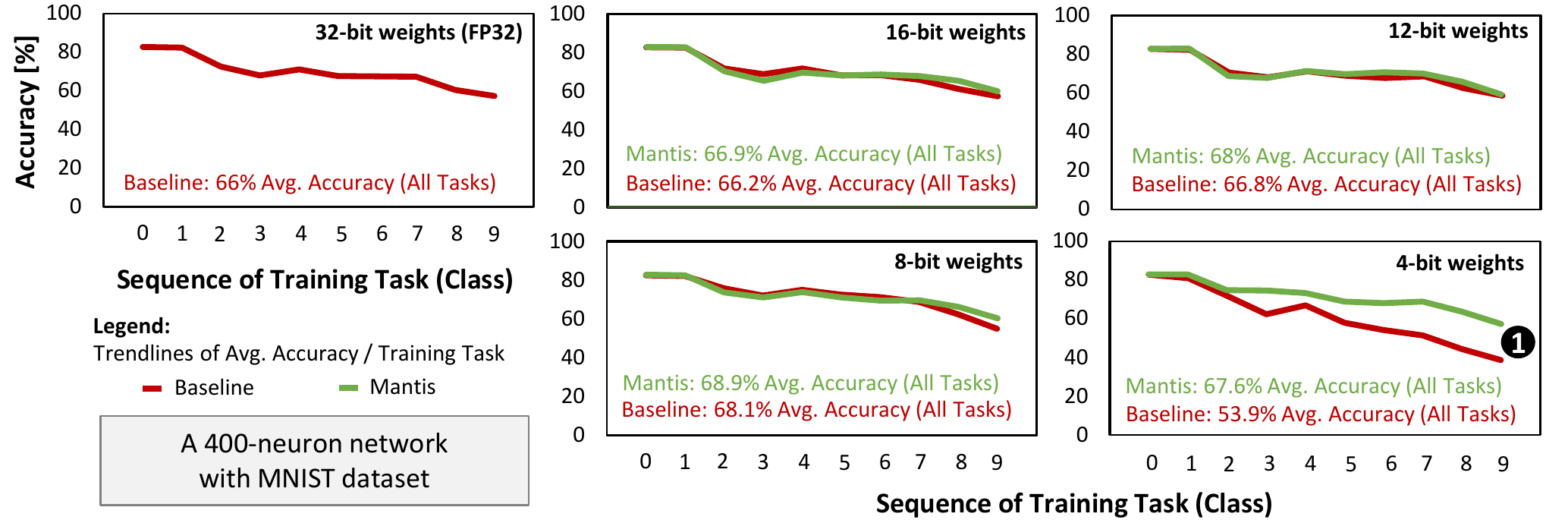}
\vspace{-0.3cm}
\caption{The accuracy profiles of a 400-neuron SNN on MNIST dataset across different levels of weight precision and across different designs (i.e., baseline and Mantis) considering dynamic environments. Note that similar trends of results are observed for GPGPU and embedded GPUs.}
\label{Fig_Results_Accuracy}
\vspace{-0.6cm}
\end{figure*}

Fig.~\ref{Fig_Results_Accuracy} presents the experimental results for SNN accuracy in dynamic environments.
These results show that our Mantis methodology maintains the accuracy over the baseline with 32-bit weights (FP32), and even improves the accuracy in certain cases, across different weight precision levels. 
For instance, label-\circledB{1} indicates that the baseline model with 4-bit weights suffers from a significant accuracy degradation (i.e., 53.9\% accuracy) as compared to its 32-bit version (i.e., 66\% accuracy).
On the other hand, Mantis successfully improves the accuracy of the SNN model with 4-bit weights (i.e., 67.6\% accuracy), which is slightly higher than the baseline (FP32). 
This accuracy improvement can be associated with several reasons. 
First, the learning enhancements in Mantis effectively perform weight updates by leveraging adaptive learning rates and balanced threshold potential to learn new input features.
Second, Mantis effectively performs weight decay to help the network gracefully remove the old and insignificant weight values, thereby providing space for learning new features.
These results highlight that our Mantis methodology provides effective learning enhancements for mobile agents, so that these agents can perform efficient online training at run time to keep the knowledge updated over their lifespan while minimizing catastrophic forgetting. 

%%%%%%%%%
\subsection{Memory Footprint Reduction}
\label{Sec_Results_Memory}

\begin{figure}[t]
\centering
\includegraphics[width=0.78\linewidth]{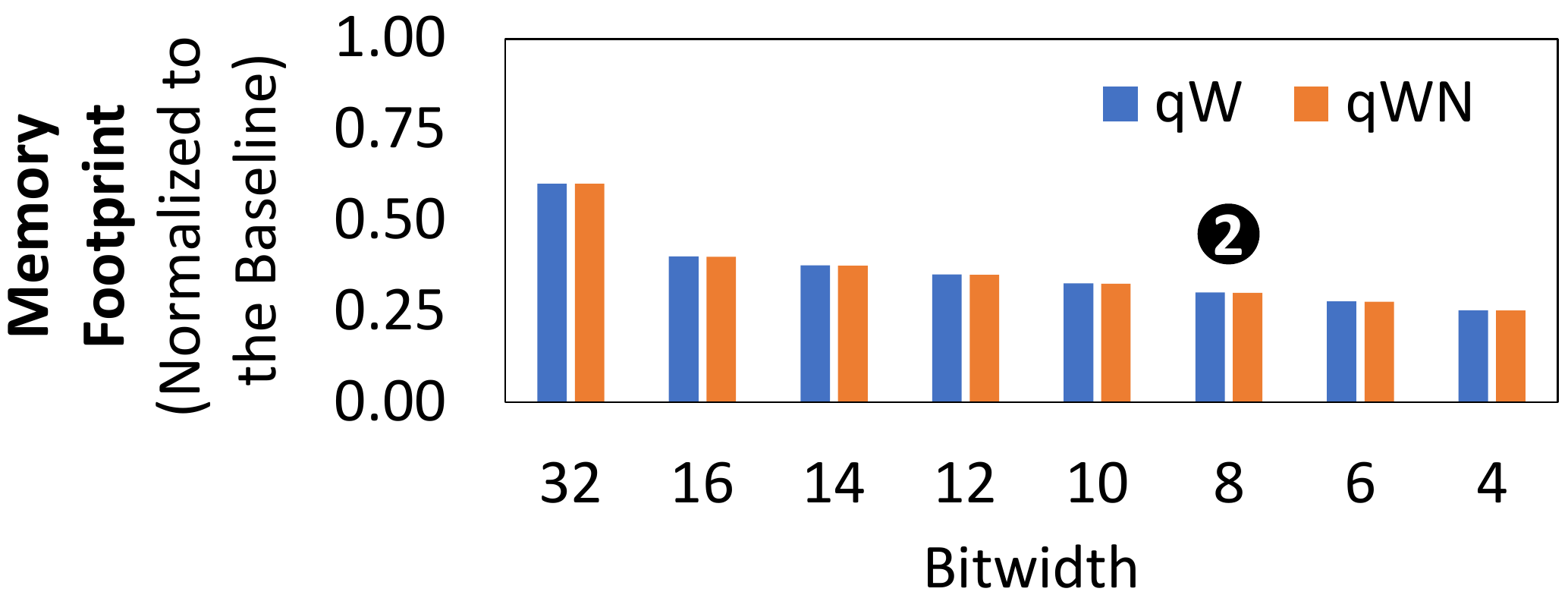}
\vspace{-0.3cm}
\caption{Memory footprint of our Mantis across different precision levels and quantized parameters, i.e., quantized weights (qW), quantized weights and neuron parameters (qWN).}
\label{Fig_Results_Memory}
\vspace{-0.3cm}
\end{figure}

Fig.~\ref{Fig_Results_Memory} presents the experimental results for memory footprint.
These results show that our Mantis effectively reduces the size of SNN models, i.e., by up to 3.3x and 3.32x memory savings for 8-bit weights (qW) and for 8-bit weights and neuron parameters (qWN), respectively; see label-\circledB{2}. 
These savings mainly come from the elimination of the inhibitory neurons and the reduction of weight precision, while additional savings come from the quantization of neuron parameters.
Note, the impact of quantized neuron parameters on the memory footprint depends on the network architecture since different architectures have a different number of weights and neuron parameters. 
For instance, a higher number of neurons leads to more neuron parameters that need to be stored in memory, hence having a higher impact on memory footprint.

%%%%%%%%%
\subsection{Reduction of Energy Consumption}
\label{Sec_Results_Energy}

\begin{figure}[hbtp]
\centering
\includegraphics[width=\linewidth]{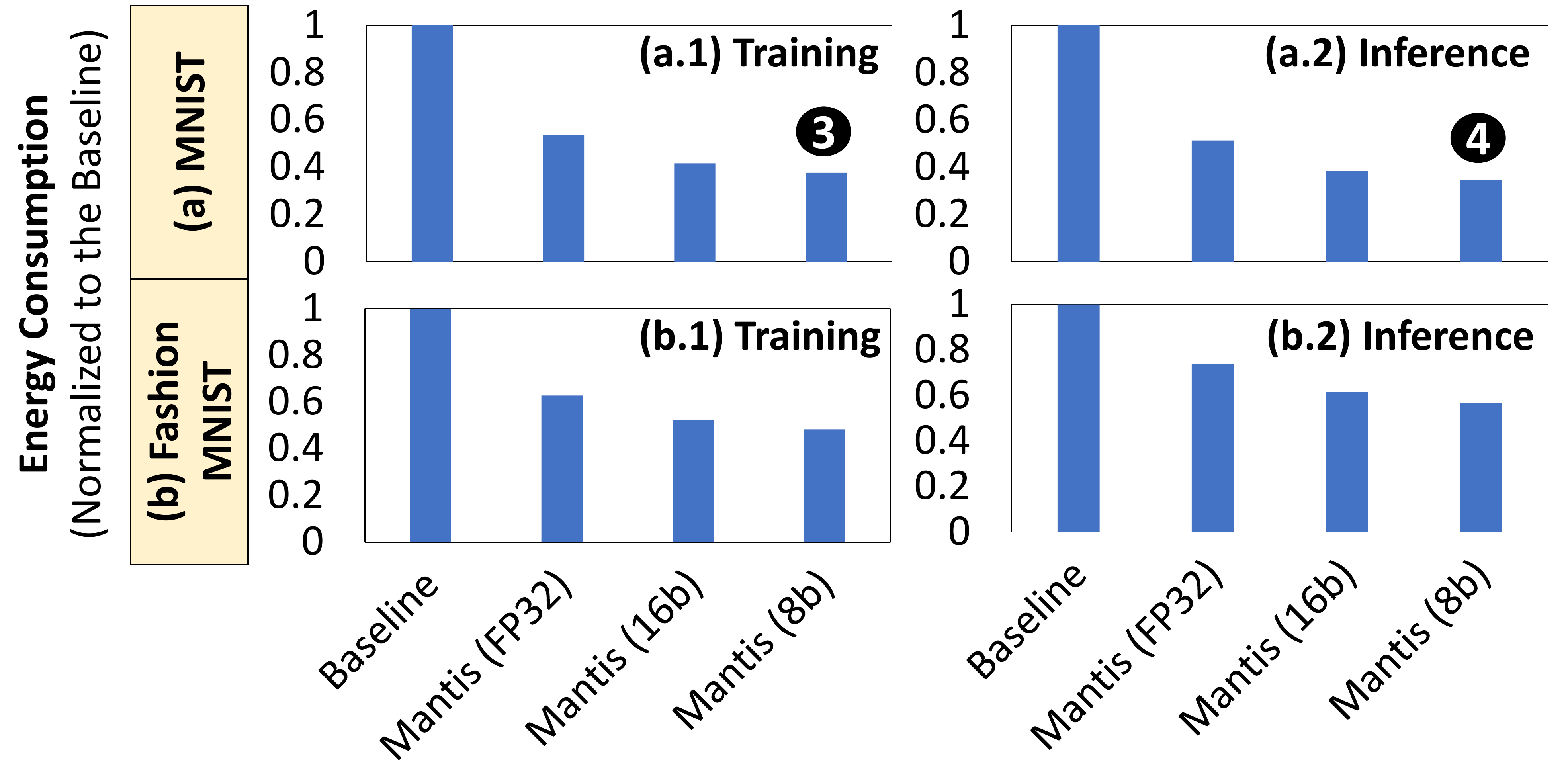}
\vspace{-0.9cm}
\caption{Energy consumption of SNN processing in training and inference for (a) the MNIST and (b) the Fashion MNIST. Similar trends are observed for GPGPU and embedded GPUs.}
\label{Fig_Results_Energy}
\vspace{-0.6cm}
\end{figure}

Fig.~\ref{Fig_Results_Energy} presents the experimental results of energy consumption of SNN processing.
These results show that our Mantis effectively reduces the energy consumption of SNNs for both the online training and inference phases and across different datasets.
For the MNIST dataset, our Mantis optimizes the energy consumption of an SNN model with 32-bit weights (FP32) by up to 1.9x in the online training phase since it eliminates the inhibitory neuron operations.
Then, the quantization in Mantis optimizes the energy consumption even further, e.g., by up to 2.7x for the quantized model with 8-bit weights; see label-\circledB{3}.
In this manner, online training can be performed by mobile agents in an energy-efficient manner.  
In the inference phase, Mantis optimizes the energy consumption of an SNN model (FP32) by up to 1.9x, mainly due to the elimination of the inhibitory neuron operations.
Then, the quantization in Mantis decreases the energy consumption even more, e.g., by up to 2.9x for the quantized model with 8-bit weights; see label-\circledB{4}.    
We also observe similar trends of results for the Fashion MNIST datasets, as shown in Fig.~\ref{Fig_Results_Energy}b.

%%%%%%%%%%%%%%%%%%%%%%%%%%%%%%%%%%%%%%%%%%%%%%%%%%%%%%%%
\section{Conclusion}
\label{Sec_Conclusion}

We present a Mantis methodology that enables memory- and energy-efficient SNNs with efficient online learning capabilities by optimizing SNN operations and employing an enhanced STDP-based learning rule.
Furthermore, Mantis also employs a model selection algorithm to find the appropriate model for the given accuracy, memory, and energy constraints.
In this manner, autonomous mobile agents can benefit from SNN characteristics to efficiently perform their functionalities. 

%%%%%%%%%%%%%%%%%%%%%%%%%%%%%%%%%%%%%%%%%%%%%%%%%%%%%%%%

\section*{Acknowledgment}
This work was partially supported by the NYUAD Center for Artificial Intelligence and Robotics (CAIR), funded by Tamkeen under the NYUAD Research Institute Award CG010. This work was also partially supported by the project ``eDLAuto: An Automated Framework for Energy-Efficient Embedded Deep Learning in Autonomous Systems”, funded by the NYUAD Research Enhancement Fund (REF).

%%%%%%%%%%%%%%%%%%%%%%%%%%%%%%%%%%%%%%%%%%%%%%%%%%%%%%%%
\vspace{-0.1cm}

\bibliography{references}
\bibliographystyle{IEEEtran}
\end{spacing}

\end{document}